\documentclass[letterpaper,11pt,twoside]{article}
\usepackage[margin=1.2in]{geometry}
\usepackage[T1]{fontenc}
\usepackage{graphicx}

\newif\iffollowingorders
\followingordersfalse

\newif\ifcomments
\commentstrue

\usepackage{latexsym}
\usepackage{booktabs}
\usepackage{enumitem}
\usepackage{graphicx}
\usepackage{color}
\usepackage{enumitem}
\usepackage{multirow}

\usepackage{algorithm}
\usepackage[noend]{algpseudocode}

\iffollowingorders
\newcommand{\citep}[1]{\cite{#1}}
\newcommand{\citet}[1]{\cite{#1}}
\else
\usepackage[numbers,square,sort&compress]{natbib}
\fi

\iffollowingorders
\else
\usepackage{amsmath}
\usepackage{amsthm}

\newcounter{myeg}
\fi

\newcommand{\BibTeX}{B\kern-.05em{\sc i\kern-.025em b}\kern-.08em\TeX}

\usepackage{graphicx}
\usepackage{xcolor,xspace}
\definecolor{RoyalBlue}{cmyk}{1, 0.50, 0, 0}
\definecolor{ForestGreen}{cmyk}{0.864, 0.0, 0.429, 0.396}
\definecolor{DarkGreen}{rgb}{0.076, 0.379, 0.306}
\definecolor{Brown}{cmyk}{0.0,0.692,0.925,0.529}
\definecolor{LightPurple}{rgb}{0.550,0.394,0.664}
\definecolor{DarkPurple}{rgb}{0.433,0.172,0.569}
\definecolor{Orange}{rgb}{0.93,0.478,0.121}
\definecolor{MyGreen}{cmyk}{0.95, 0.05, 0.95, 0.05}
\definecolor{MyPurple1}{rgb}{0.45,0.353,0.963}
\definecolor{MyPurple2}{rgb}{0.63,0.4,0.63}
\definecolor{MyYellow}{rgb}{0.901,0.547,0.0}
\definecolor{CambridgeRed}{rgb}{0.797, 0.000, 0.001}

\usepackage{pifont}
\newcommand{\cmark}{\ding{51}}%
\newcommand{\xmark}{\ding{55}}%
\usepackage{amsfonts}

\usepackage[
    breaklinks=true
   ,colorlinks%
   ,plainpages=false
   ,hypertexnames=true
   ,naturalnames
   ,hyperindex
   ,citecolor=MyGreen 
   ,urlcolor=blue
   ,pdfauthor={}
   ,pdftitle={}
   ,pdfsubject={}
]{hyperref}

\usepackage{bm}
\usepackage{multirow}

\usepackage{nicefrac}
\usepackage{comment}

\usepackage{paralist}
\usepackage{upgreek}

\usepackage{multicol}

\usepackage{subcaption}

\usepackage{pstricks}

\usepackage{enumitem}

\usepackage{booktabs}       

\usepackage{wrapfig}
\usepackage{framed}

\usepackage{amsmath}

\newcommand{\abs}[1]{\left\lvert#1\right\rvert}

\newcommand{\set}[1]{\ensuremath{\left\{#1\right\}}\xspace}
\newcommand{\multiset}[1]{\set{\set{#1}}}

\newcommand{\expectedsub}[2]{\ensuremath{\mathbf{E}_{#1}\left[#2\right]}\xspace}

\newcommand{\XX}{\ensuremath{\mathcal{X}}\xspace} 

\newcommand{\sample}{\ensuremath{S}\xspace}

\newcommand{\multiflow}{\ensuremath{\mbox{\texttt{scikit-multiflow}}}\xspace}
\newcommand{\river}{\ensuremath{\mbox{\texttt{River}}}\xspace}

\newcommand{\empdensity}[2]{\ensuremath{\tilde{f}_{#1}\left(#2 \right)}\xspace}
\newcommand{\kde}[2]{\ensuremath{\hat{f}_{#1}\left(#2 \right)}\xspace}

\newcommand{\mygraphic}[1]{\includegraphics[height=#1]{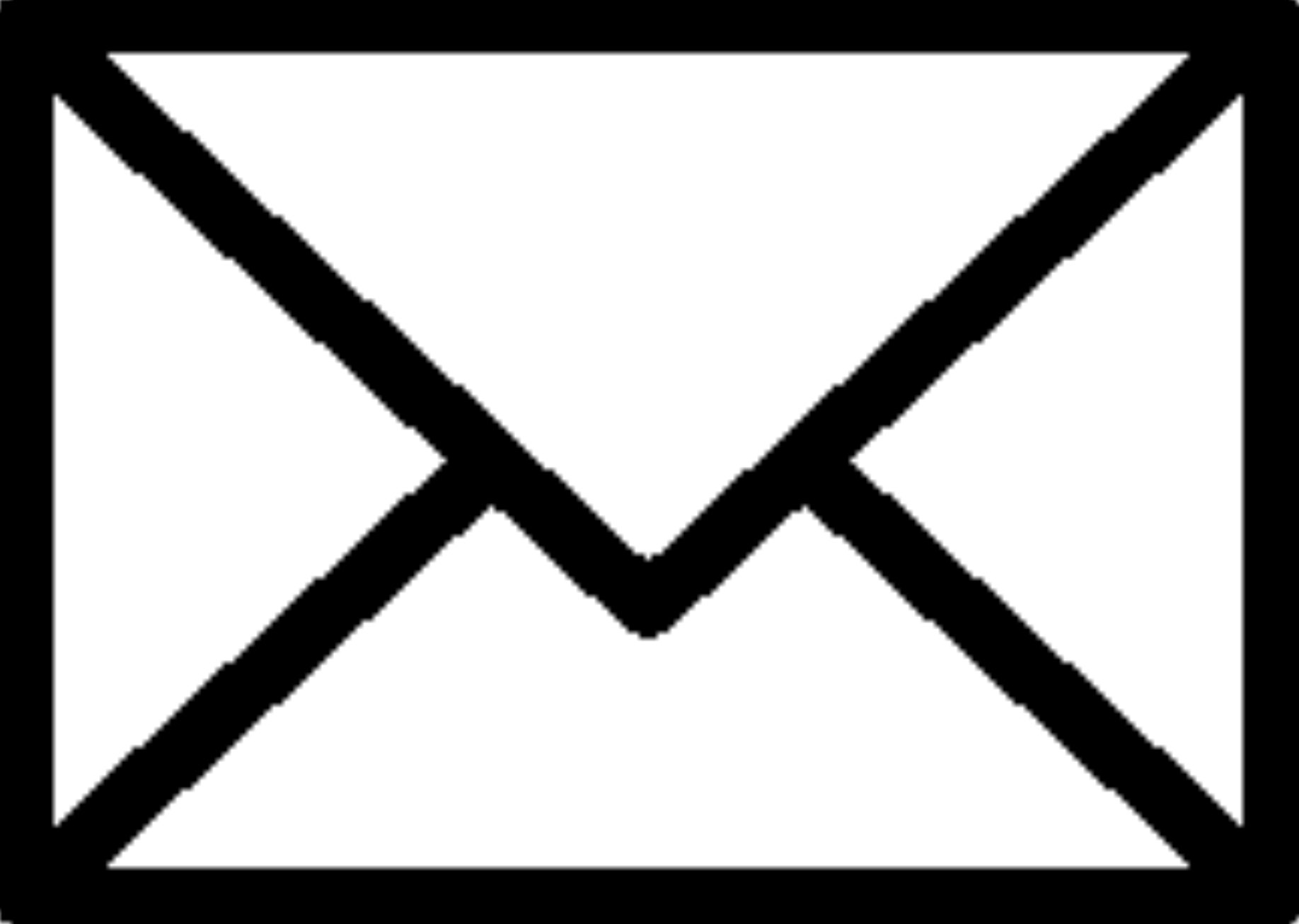}}

\begin{document}

\title{On Imbalanced Regression with Hoeffding Trees}

\author{Pantia-Marina Alchirch\textsuperscript{(\mygraphic{0.6em})}\href{https://orcid.org/0000-0002-9596-7884}{\includegraphics[height=0.35cm]{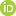}}
\and 
Dimitrios I.~Diochnos\href{https://orcid.org/0000-0002-2934-606X}{\includegraphics[height=0.35cm]{figs/author/ORCIDiD_icon16x16.png}}}
\date{University of Oklahoma\\
{\small\texttt{\{marina.alchirch,diochnos\}@ou.edu}}}

\maketitle              

\begin{abstract}
%
%
Many real-world applications generate continuous data streams for regression. 
Hoeffding trees and their variants have a long-standing tradition due to their effectiveness, either alone or as base models in broader ensembles.
Recent batch-learning work shows that \emph{kernel density estimation (KDE)} improves smoothed predictions in imbalanced regression~\citep{Yang2021DelvingID}, while \emph{hierarchical shrinkage (HS)} provides post-hoc regularization for decision trees without modifying their structure~\citep{DT:Regularization:HierarchicalShrinkage}.
We extend KDE to streaming settings via a telescoping formulation and integrate HS into incremental decision trees. Empirical evaluation on standard online regression benchmarks shows that KDE consistently improves early-stream performance, whereas HS provides limited gains. Our implementation is publicly available at:~\href{https://github.com/marinaAlchirch/DSFA_2026}{https://github.com/marinaAlchirch/DSFA\_2026}.

\end{abstract}

%


\section{Introduction}\label{sec:intro}
We live in a world where sensors continuously monitor the environment and our activities, generating data with the potential to transform our lives. In this context, machine learning and data mining algorithms are invaluable, as they uncover patterns that reveal important phenomena. Mining such continuous flows of information gives rise to \emph{online} machine learning and \emph{data stream} mining and has enabled applications such as credit scoring~\citep{Applications:CreditScore}, fraud detection~\citep{Applications:Fraud}, human activity recognition~\citep{Applications:HAR}, clinical decision support~\citep{Applications:Patients}, energy pricing~\citep{Applications:EnergyPricing}, and even reducing the energy footprint of stream mining itself~\citep{Applications:GreenAI}.

At the same time, 
working with \emph{imbalanced data} is a constant nuisance for 
machine learning and data mining applications.
Imbalanced data arise when the labels associated with the observations that are collected, are skewed favoring certain categories 
or certain ranges of values.
Research on imbalanced data has focused primarily on classification problems; e.g., 
anomaly detection~\citep{Imbalanced::AnomalyDetection}, 
e-commerce~\citep{Imbalanced::Ecommerce}, 
infants' survival~\citep{Imbalanced::Gourdeau}, 
and many others.
However, investigating imbalanced data makes a lot of sense in regression tasks as well. For example, one may want to predict the age of an individual by having access to their facial image~\citep{Imbalanced:DIR:AgeApplication},
or 
predict weather phenomena such as hail size or wind intensity~\citep{Imbalanced::McGovern:1}.

Our work focuses at the intersection of the above two fields.  We investigate popular methods that generate decision trees incrementally and have been developed with ideas for mining data streams and combine such mechanisms with recent advances developed for batch learning: \textit{(i)} regularization of decision trees using hierarchical shrinkage~\citep{DT:Regularization:HierarchicalShrinkage}, and \textit{(ii)} approximating skewed distributions using kernel density estimation (KDE)~\citep{Yang2021DelvingID}.
In this work we are interested in imbalanced data; not in concept drift.
Below we summarize our contributions.

\paragraph{Contributions.}
First, we implement hierarchical shrinkage on incremental decision trees that are available in the \multiflow library~\citep{Library:Multiflow}.
Second, we revisit KDE and make it applicable to mining algorithms that operate on streams.
Third, by combining the above techniques we achieve improved performance of incremental decision tree algorithms on imbalanced regression tasks for streams.
Our main finding is that KDE helps with the improved predictions a lot, whereas hierarchical shrinkage provides minimal gains.
We then integrate KDE with the \river library~\citep{Library:River} and verify that KDE improves predictions in the same tree models tested in \multiflow, as well as in additional ones that were not tested or were not available in \multiflow.
Our code is publicly available for anyone to work with: \href{https://github.com/marinaAlchirch/DSFA_2026}{https://github.com/marinaAlchirch/DSFA\_2026}.

\section{Related Work}\label{sec:related_work}
Incremental decision trees have a long line of research going back to at least the 80's~\citep{DT:Incremental:ID4,DT:Incremental:ID5}. 
Perhaps the most important incremental decision tree learning algorithm is that of Hoeffding trees~\citep{DT:Incremental:VFDT} and their variants. 
Hoeffding trees use Hoeffding's inequality as a heuristic in order to decide when the difference in information gain (or some other relevant criterion) between the best-split attribute and the second-best split attribute is sufficiently high to justify a split. 
Several extensions of the Hoeffding trees have been developed~\citep{DT:Incremental:HAT,DT:Incremental:VFDTc,Tree:SGT,DT:Incremental:CVFDT,HT:FIMT-FIRT,Tree:iSOUP:RiverCitation}, primarily to deal with concept drift, but attention has also been paid for parallelization~\citep{DT:Incremental:SPDT}, reduced complexity~\citep{HT:Regularized}, or improved classification accuracy~\citep{DT:Incremental:ExtremelyFastDT}.

In parallel, a tremendous wealth of methods has been developed for mining imbalanced data.
A well-known method for dealing with class imbalance is Synthetic Minority Oversampling Technique (SMOTE) in which oversampling of synthetic data is performed~\citep{Imbalanced::Smote,Imbalanced::SmoteRegression}.
However, many more methods have been developed, including 
informed under-sampling the majority class~\citep{Imbalanced::UndersamplingInformed}, sampling based on clusters~\citep{Imbalanced::Clustering}, re-weighting~\citep{Imblanced::neurips2017}, modifying the margin of the separation boundary~\citep{Imbalanced::neurips2019}, or more  theoretical approaches~\citep{Imbalanced:ThrowAway,Imbalanced:margin,Diochnos:PACI} including more general scenarios~\citep{Imbalanced:Mansour}.
In addition, a recent line of work is using \emph{kernel density estimation (KDE)} for smoothing the predicted labels on learned models~\citep{Imbalanced:DIR:AgeApplication,Yang2021DelvingID}, but has been applied only in batch settings where all the information is available ahead of time.
Inspired by this approach we smooth the predicted labels using sketches of the distribution from small sequences of the stream examples.

Finally, a useful way for regularizing decision trees is that of \emph{pruning}~\citep{Book:CART,Book:C45}. 
However,
pruning is rarely used for incremental trees since it requires constructing a full tree and then pruning branches in a costly post-processing step.
A more recent alternative for regularization is \emph{hierarchical shrinkage (HS)}~\citep{DT:Regularization:HierarchicalShrinkage}, where all nodes along the root-to-leaf path contribute to the final prediction. HS is a post-hoc regularization method that avoids building a complete tree or performing expensive pruning operations, and can be applied to incremental trees provided sufficient streaming statistics are maintained.
To our knowledge, we are the first to integrate HS into incremental decision trees and evaluate its impact on predictive accuracy, coupled with KDE-based smoothing for imbalanced domains.

\section{Preliminaries}\label{sec:preliminaries}


We use \emph{multisets} for collections of objects that may contain multiple copies 
of the same object; 
e.g.,~$\sample = \multiset{x_1, x_3, x_1}$.
For a compact representation denoting multiplicities, we may write 
$\sample = \multiset{x_1, x_3, x_1} = \multiset{(x_1, 2), (x_3, 1)}$.
The \emph{cardinality} $\abs{\sample}$ of a multiset \sample is the total number of elements in \sample 
(including
multiplicities); e.g., 
$\abs{\sample} = 3$ for the multiset \sample given above.
%
Using multisets we have the ability to approximate \emph{probability density functions (pdfs)} that govern certain domains by keeping track of their \emph{empirical densities}.  Furthermore, in these cases we will 
explicitly indicate that certain elements do not appear at all, using 0 for their multiplicities. For example, in an instance space $\XX = \set{x_1, x_2, x_3, x_4}$ using the collection of instances $\sample = \multiset{(x_1, 2), (x_3, 1)}$, we have the empirical density $\empdensity{\sample}{\XX} = \set{(x_1, 2), (x_2, 0), (x_3, 1), (x_4, 0)}$.
With $m_{\mathcal{Z}}(z)$ we denote the multiplicity of $z$ in $\mathcal{Z}$; e.g., $m_{\empdensity{\sample}{\XX}}(x_1) = 2$ in the previous example.

\paragraph{Kernel Density Estimation (KDE).}
Kernels provide an easy way for similarity comparisons between instances. By storing instances in certain windows we can provide smooth estimates on imbalanced distributions for label values.
In general, \emph{kernel density estimation (KDE)}~\citep{parzen1962estimation} using some arbitrary kernel $K$, points $z_i\in\sample$ from some multiset \sample, 
and a query point $q$,
is accomplished via: 
\begin{equation}\label{eq:kernel_general}
\kde{\sample,K,h}{q}
 = \frac{1}{\abs{\sample}}\sum_{(z_i, m_i)\in\sample} m_i\cdot K_h(q-z_i) 
= 
\frac{1}{\abs{\sample}h}\sum_{(z_i, m_i)\in\sample} m_i\cdot K\left(\frac{q-z_i}{h}\right)\,,
\end{equation}
where $h$ is the \emph{bandwidth} and its role is to provide some scaling on the use of the kernel that is used in (\ref{eq:kernel_general}).
Small values of $h$ make the estimate sensitive to individual data points, whereas large values of $h$ provide a smoother estimate, by spreading the influence of each data point over a wider region of the dataset. 
%
We work with the Gaussian and the Epanechnikov kernel.
The approximation on the labels as explained above is called LDS in~\citep{Yang2021DelvingID}.

Furthermore, we may use \emph{binning} to map individual label values to broader intervals.  This is standard technique; e.g.,~\cite{Yang2021DelvingID}. 
Let $m$ and $M$ denote the minimum and maximum label values (assumed known), and let $r$ be the bin range. We generate $j^\star = \lfloor (M-m)/r \rfloor + 1$ bins, where 
$b_j = [m + jr, m+(j+1)r)$ for $j\in\set{0,1,\ldots,j^\star}$.
Since~(\ref{eq:kernel_general}) ultimately computes an average, it admits a telescoping update that relies on the previous average and the newest observation; which is particularly convenient in streaming settings. 
Using $n$ in the first subscript to denote the fact that we have seen $n$ examples so far from the stream, we have:
\begin{equation}\label{eq:telescoping_kde}
    \hat{f}_{n,K,h}(q) = \hat{f}_{n-1,K,h}(q) + \frac{1}{n}\left(\frac{1}{h}K\left(\frac{q-z_n}{h}\right) - \hat{f}_{n-1,K,h}(q)\right)\,.
\end{equation}
Thus, we are able to implement KDE incrementally; Algorithm~\ref{alg:incr_kde} has the details.

\begin{algorithm}[t]
    \caption{Incremental KDE}\label{alg:incr_kde}
    \begin{algorithmic}[1]
        \Require Window $W = \{(x_1, y_1), (x_2, y_2), \dots, (x_m, y_m) \}$, bandwidth $h$, kernel $K$, and a structure of $B$ groups of bins.
        \Ensure Smoothed weights $\hat{f}_{n, K, h}(b)$ for each bin $b \in B$, where $n$ is number of examples that we have seen in the stream so far.
        
        \For{target label $y_i \in W$}
        \State Find bin $b = \lfloor (y_i - m)/r \rfloor$ for target label $y_i$
        \State Compute $\hat{f}_{n+i, K, h}(b)$ as shown in~(\ref{eq:telescoping_kde})
        \EndFor
    \\    
    \Return$\{\kde{n+m, K, h}{b}\}$, for all b $\in B$
    \end{algorithmic}
    \end{algorithm}

\paragraph{Hoeffding Trees and Variants.}
The most well-known incremental decision tree algorithm is the Hoeffding Tree (HT)~\citep{DT:Incremental:VFDT} and its variants. This class of algorithms uses Hoeffding’s inequality to decide when to split a node. Letting $R$ be the range of the split evaluation function (e.g., $R=1$ for information gain in binary classification), when the difference between the best and second-best attributes exceeds
$\epsilon = \sqrt{\frac{R^2\ln(1/\delta)}{2n}}$,
then, with confidence at least $1-\delta$, the best attribute is selected for splitting.
For our work we use the \multiflow~\citep{Library:Multiflow} and \river~\citep{Library:River} libraries. 
The \multiflow library has an implementation of HT and of the \emph{Hoeffding Adaptive Tree (HAT)}~\citep{DT:Incremental:HAT}. HAT employs ADWIN~\cite{ADWIN} to detect potential concept drift.
Similar to \multiflow the \river library has an implementation of HT and HAT.  Sticking to regression, \river also has an implementation of iSOUP~\citep{Tree:iSOUP:RiverCitation} and SGT~\citep{Tree:SGT}.

\paragraph{Regularization of Decision Trees Using Hierarchical Shrinkage.} 
Hierarchical shrinkage (HS)~\citep{DT:Regularization:HierarchicalShrinkage} is a recent regularization technique that does not alter the structure of the learned tree.
Letting $N(t)$ be the number of samples in a node $t$ of a decision tree $DT$ and $\expectedsub{t}{y}$ to be the mean response in node $t$, 
the prediction on an instance $x$ is
$DT(x) = \expectedsub{t_0}{y} + \sum_{l=1}^L\left(\expectedsub{t_l}{y} - \expectedsub{t_{l-1}}{y}\right)$.
HS modifies this 
to
\begin{equation}
DT_\lambda(x) = \expectedsub{t_0}{y} + \sum_{l=1}^L\frac{\expectedsub{t_l}{y} - \expectedsub{t_{l-1}}{y}}{1 + \lambda/N(t_{l-1})}\,,
\end{equation}
for some hyper-parameter $\lambda$ that performs the regularization. 
Thus, the nodes in the path from root to leaf to have an amount of say in the final prediction.
%
%

\section{The Online Learning Process That We Study}\label{sec:implementation}
We use a variation of the \emph{Follow-the-Leader (FTL)} algorithm~\cite{FollowTheLeader} 
for tuning.
In FTL, multiple models run in parallel and, at each time step, prediction follows the model that performs best (e.g., lowest cumulative loss) at that point in time. 
To ease the complexity and memory requirements 
we periodically allocate portions of the stream for such tuning and, after each phase, we select the hyper-parameters that performed best in the most recent tuning period. 
This procedure is summarized in Algorithm~\ref{alg:online_ftl} and explained further below.
%

Tuning and learning on the stream is governed by two parameters: a parameter $s_t$ controlling the duration of each tuning phase, and a parameter $s_t'$ specifying how many examples are processed in a predict-then-train manner between tuning phases. 
The model used for prediction is denoted by $M$; it is selected as the best-performing model from a pool of models $T$. In Line 3, we initialize $T$ with all instantiations of the base learner (e.g., Hoeffding Tree) across all combinations of hyper-parameter values. 
Hence, during tuning we effectively perform a grid search, where each grid point corresponds to the same base model with different hyper-parameters. Unlike standard grid search, 
all models in $T$ are trained in parallel on the stream. For example, one configuration may consist of a Hoeffding Tree combined with KDE using a Gaussian kernel with bandwidth $h$, bin range $r$, and no hierarchical shrinkage ($\lambda=0$).

\begin{algorithm}[t]
\caption{Online-Learning Process \& Follow-the-Leader Tuning}\label{alg:online_ftl}
\begin{algorithmic}[1]
\Require Stream $S = \{(x_1, y_1), (x_2, y_2), \dots, (x_n, y_n), \dots \}$ of labeled examples, a set $P$ of parameters, a tuning window size $s_t$, and no-tuning window size $s_t'$.
\State Initialize a model M;
\State M.parameters $\gets \emptyset$; 
\State Initialize all different models $T$ based on $P$;
\State $i \gets 1$;
\State $\mathrm{tuneStart}$ $\gets 1$;
\State $\mathrm{tuneEnd} \gets s_t$;

\For{every example $(x_i, y_i)$ in stream $S$}

\If{$i \geq \mathrm{tuneStart}$ and $i \le \mathrm{tuneEnd}$}
\State $j \gets 1$; 
\State minError $\gets \infty$; 
\State bestModel $\gets \emptyset$;
\While{$j \le s_t$}
\State Predict-then-train with every model $t \in T$;
\If{minError $\geq$ t.error}
\State minError $\gets$ t.error;
\State bestModel $\gets$ t;
\EndIf
\State $j \gets j + 1$;
\EndWhile
\State $i \gets i + s_t$; 
\State $\mathrm{tuneStart} \gets \mathrm{tuneStart} + s_t + s_t'$; 
\State $\mathrm{tuneEnd} \gets \mathrm{tuneStart} + s_t - 1$;
\State M.setParameters(bestModel.parameters);
\Else
\State Predict-then-Train on example $(x_i, y_i)$ with M;
\State $i \gets i + 1$;
\EndIf 
\EndFor
\end{algorithmic}
\end{algorithm}

\section{Experiments}\label{sec:experiments}

\begin{figure}[!ht]
\centering
\begin{subfigure}[b]{0.48\columnwidth}
\centering
\includegraphics[width=\columnwidth]{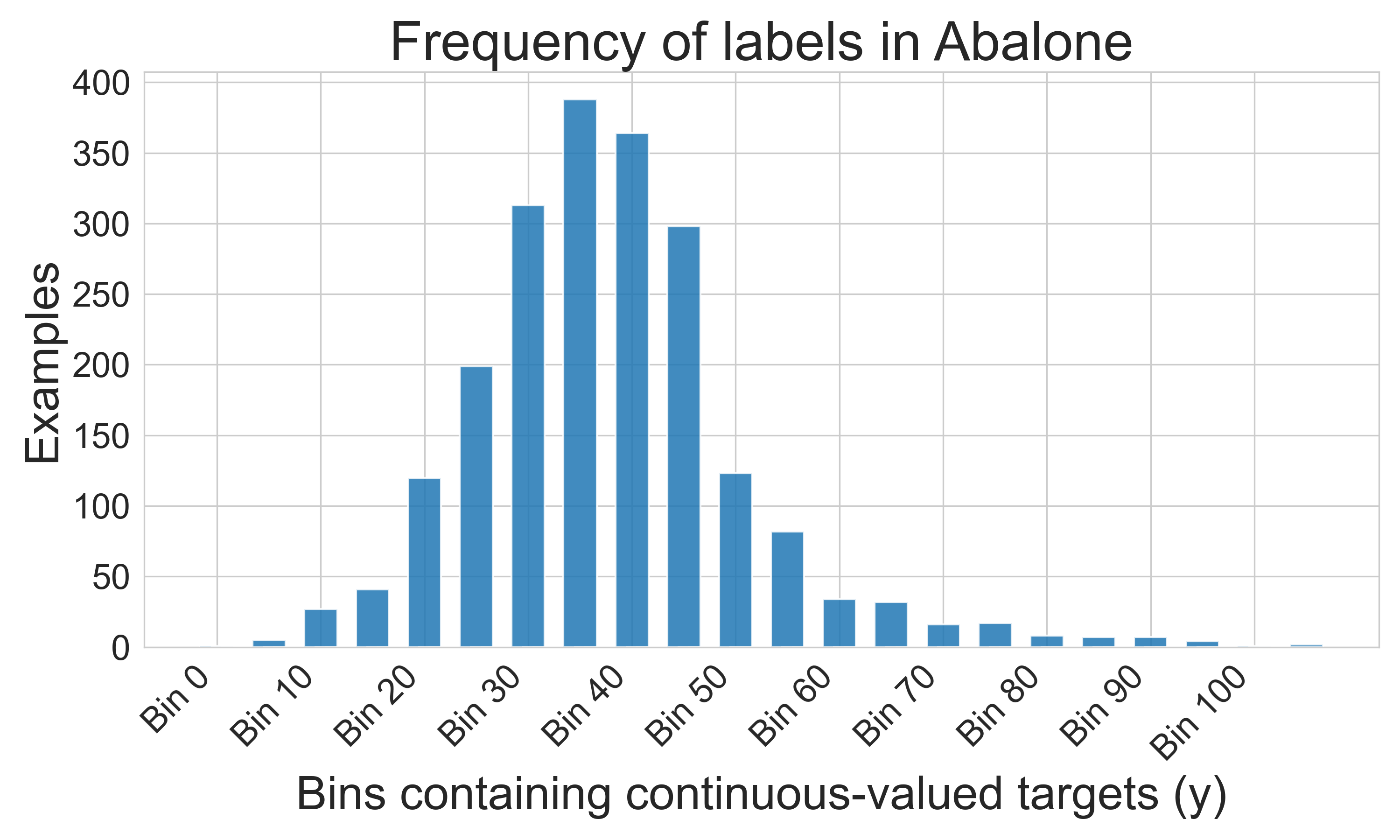}
\caption{Abalone}
\label{fig:abalone_imbalance}
\end{subfigure}
\hspace{\fill}
\begin{subfigure}[b]{0.48\columnwidth}
\centering
\includegraphics[width=\columnwidth]{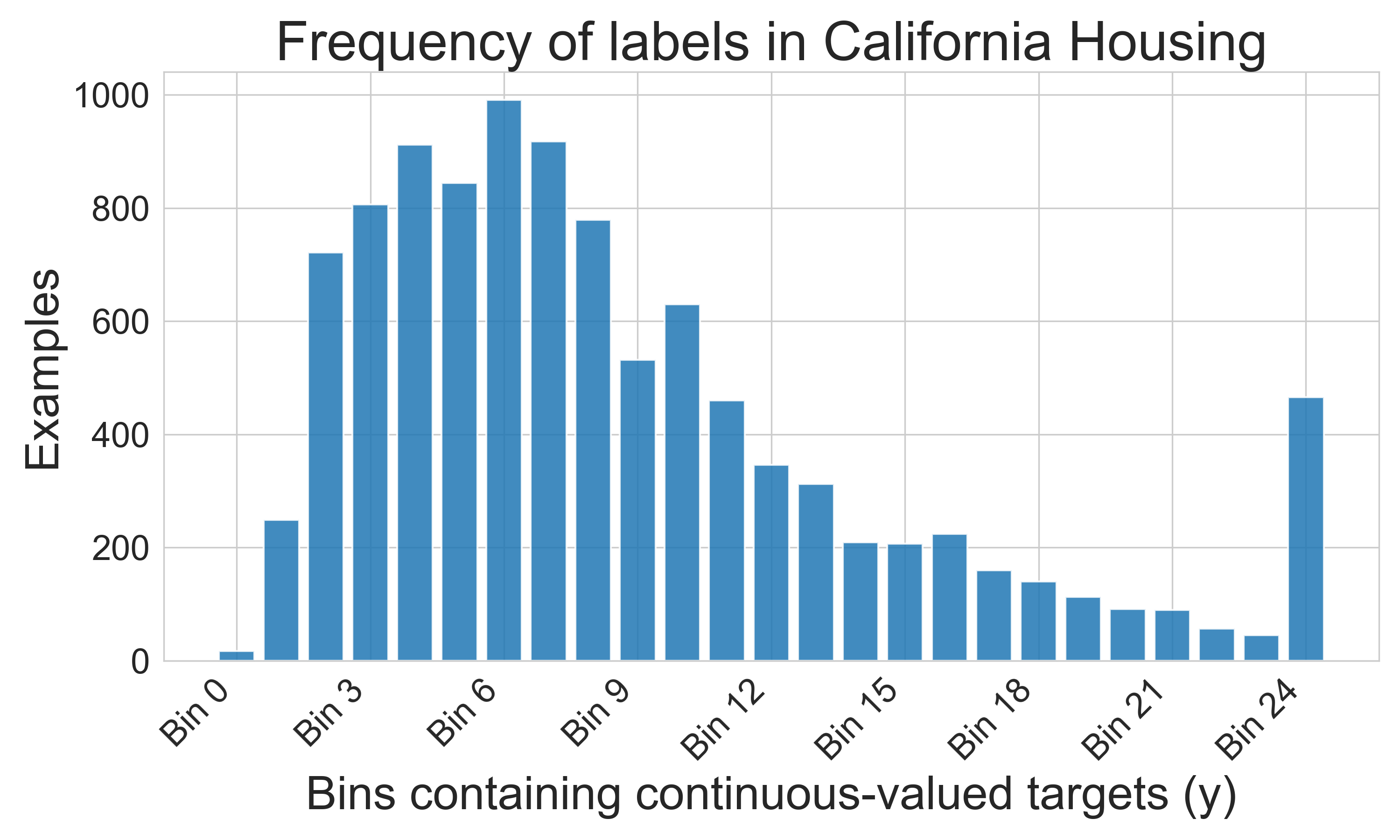}
\caption{California Housing}
\label{fig:cali_imbalance}
\end{subfigure}
\\
\begin{subfigure}[b]{0.48\columnwidth}
\centering
\includegraphics[width=\columnwidth]{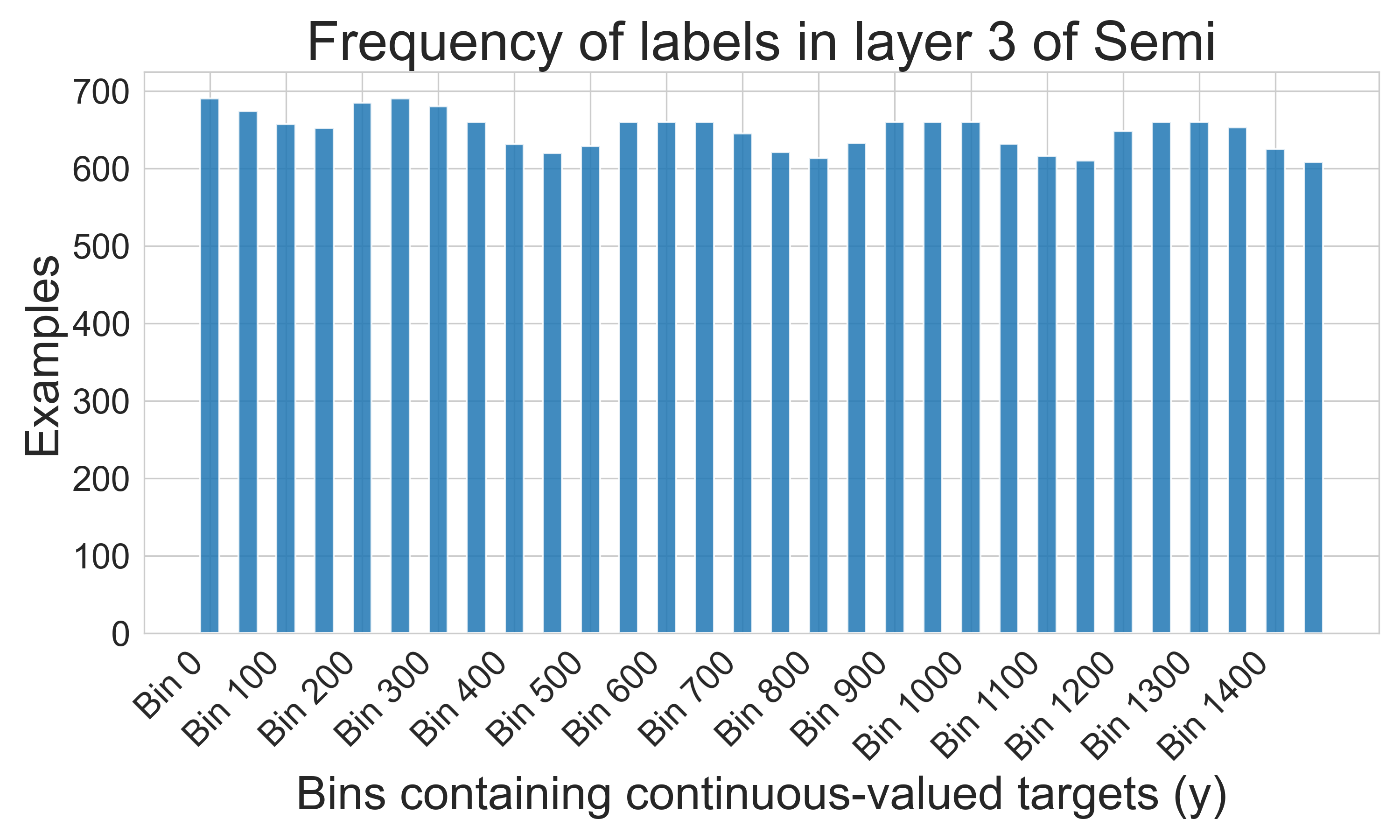}
\caption{Film Thickness of layer 3}
\label{fig:layer3_imbalance}
\end{subfigure}
\hspace{\fill}
\begin{subfigure}[b]{0.48\columnwidth}
\centering
\includegraphics[width=\columnwidth]{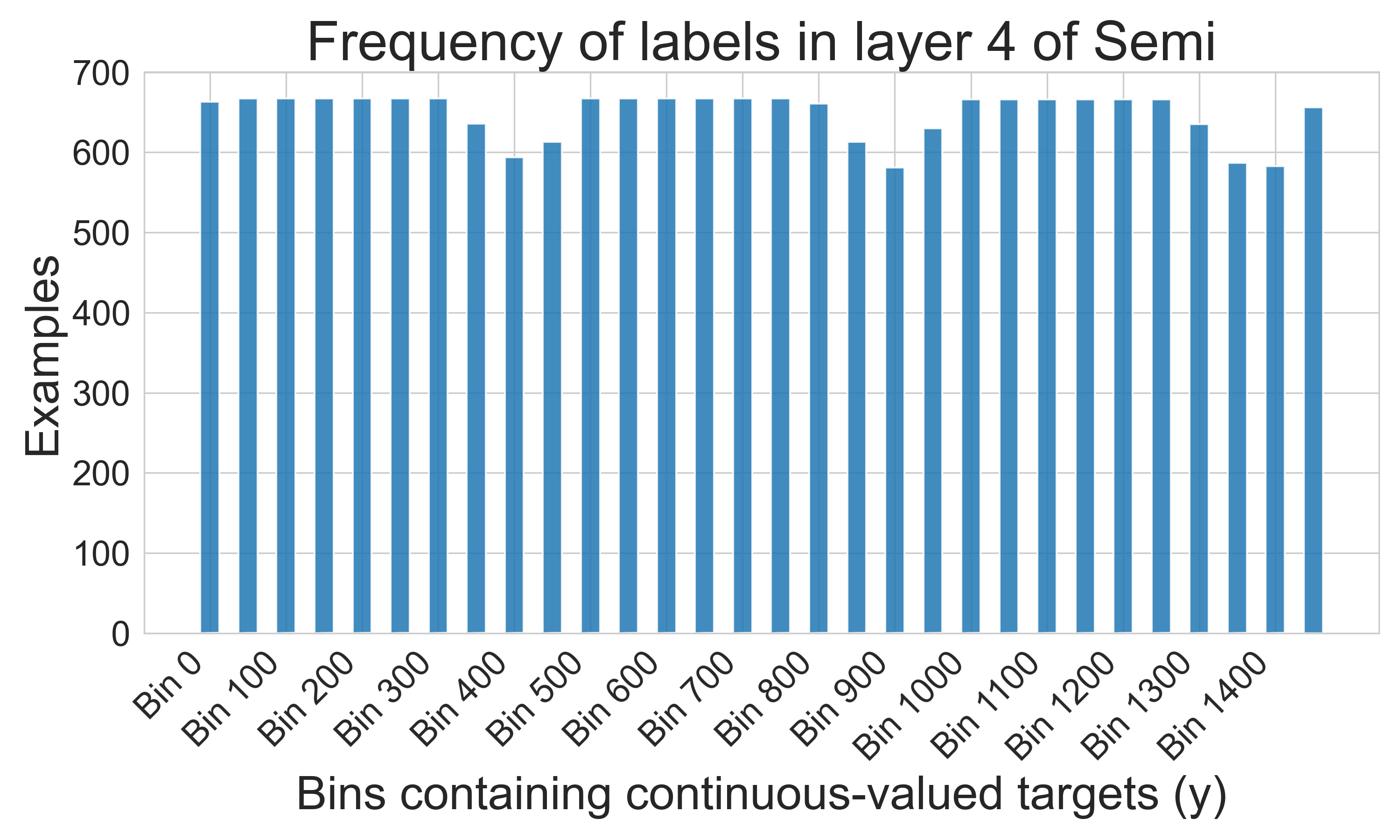}
\caption{Film Thickness of layer 4}
\label{fig:layer4_imbalance}
\end{subfigure}
\\
\begin{subfigure}[b]{0.48\columnwidth}
\centering
\includegraphics[width=\columnwidth]{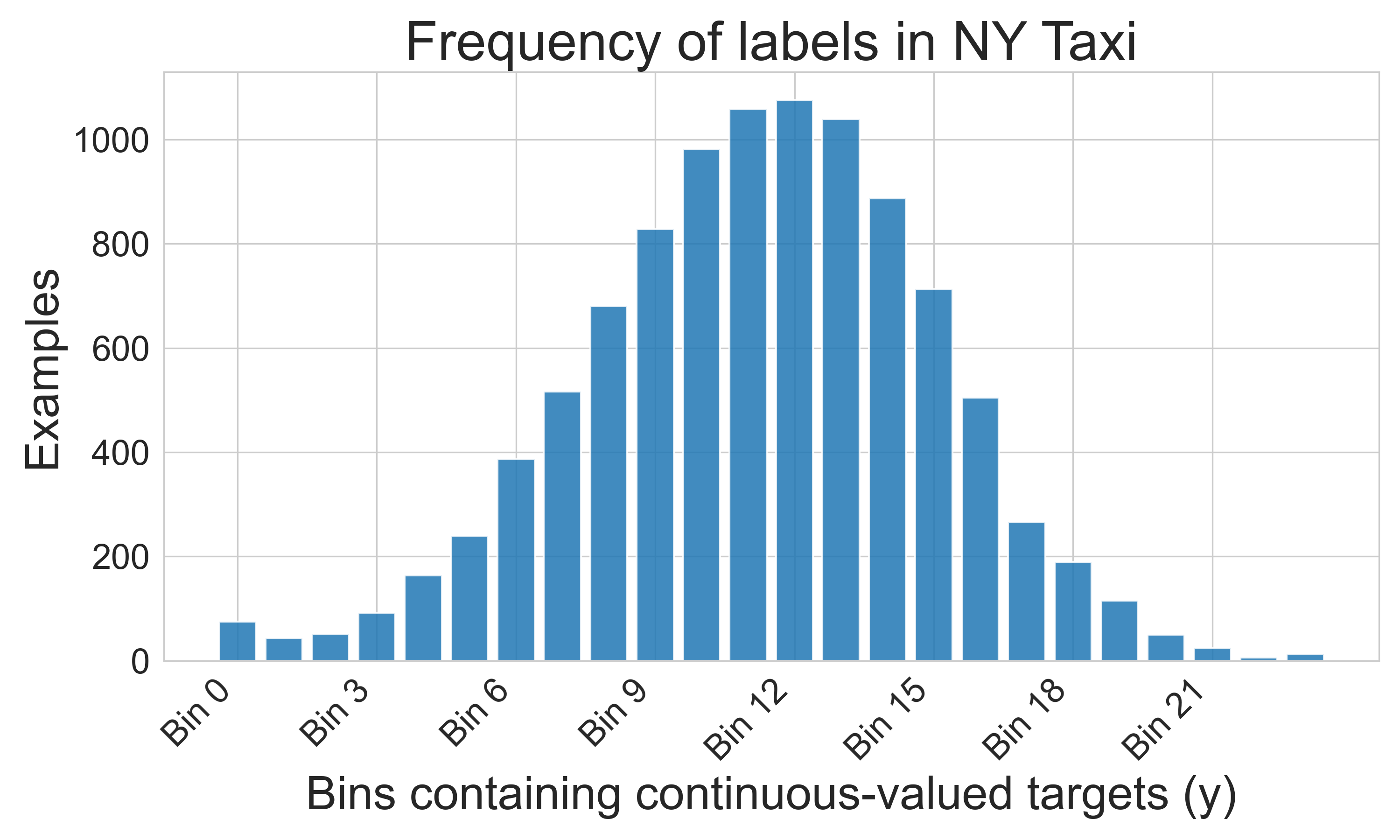}
\caption{NY Taxi}
\label{fig:nytaxi_imbalance}
\end{subfigure}
\hspace{\fill}
\begin{subfigure}[b]{0.48\columnwidth}
\centering
\includegraphics[width=\columnwidth]{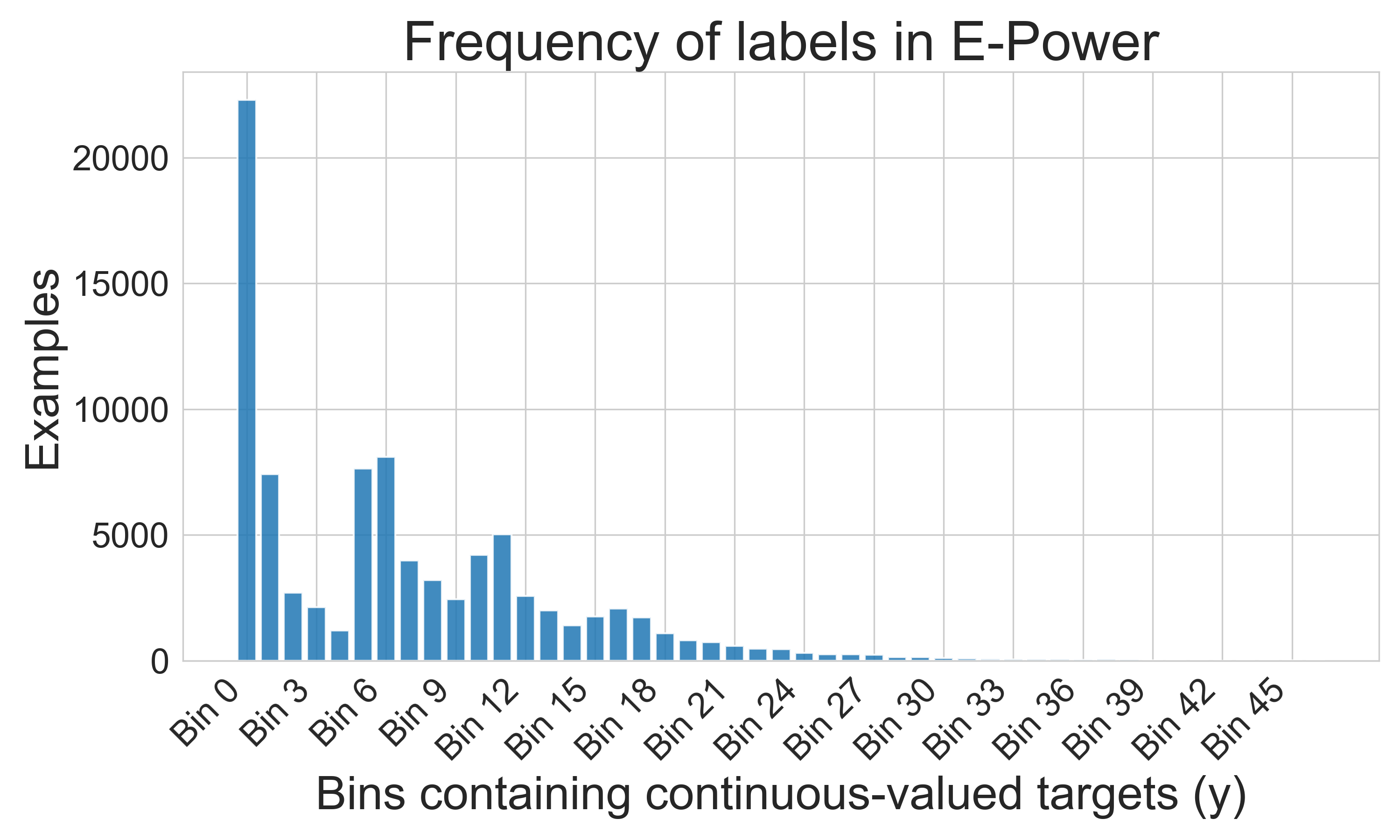}
\caption{E-Power}
\label{fig:epower_imbalance}
\end{subfigure}
\caption{Frequency of labels in predict-then-train data using bins' range $r=0.2$.}
\label{fig:imbalance_of_targets_in_datasets}
\end{figure}

\paragraph{Datasets.}
We conduct experiments on Abalone~\citep{Data:UCIrepository}, California Housing (California)~\citep{KELLEYPACE1997291CaliforniaData}, Electric Power Consumption (E-Power)~\citep{Data:UCIrepository}, New York Taxi (NY Taxi)~\citep{Data:NY:Taxi}, and Film Thickness in semiconductors (Semi)~\citep{Data:Dacon:Semiconductor}; see Table~\ref{tbl:datasets} for dataset characteristics.
\begin{table}[t]
    \centering
    \caption{Datasets and potential preprocessing for the experiments. Online tuning was tried in all datasets except for Semi due to time constraints.}
    \label{tbl:datasets}
    \begin{tabular}{cccccc}
    \toprule
         &  &  & \textbf{Predict-then-train} & & \\    
         &  &  & \textbf{set size used in} & \textbf{Outlier} & \textbf{Constructed} \\    
        \textbf{Dataset} & \textbf{\#Features} & \textbf{\#Examples} & \textbf{Algorithm~\ref{alg:online_ftl}} & \textbf{removal} & \textbf{feature} \\
    \midrule
        Abalone & 8 & 4177 & 2089 & \textcolor{magenta}{\xmark} & \textcolor{magenta}{\xmark} \\
        \cmidrule{1-6}
        Semi & 229 & 19452 & - & \textcolor{MyGreen}{\cmark}  & \textcolor{magenta}{\xmark} \\
        California & 8 & 20635 & 10319 & \textcolor{MyGreen}{\cmark} & \textcolor{magenta}{\xmark} \\
        \cmidrule{1-6}
        NY Taxi & 10 & 19997 & 10001 & \textcolor{MyGreen}{\cmark} & \textcolor{MyGreen}{\cmark} \\
        E-Power & 8 & 99992 & 87992 & \textcolor{MyGreen}{\cmark} & \textcolor{MyGreen}{\cmark} \\
    \bottomrule
    \end{tabular}
\end{table}
Abalone is small and we are using it ``as-is''.
For the remaining datasets we remove outliers, and for NY Taxi and E-Power we additionally construct standard features (e.g., Manhattan distance for NY Taxi; hour and weekday for E-Power).
For California Housing, we use 20,635 examples after removing 5 extreme instances. The NY Taxi dataset (currently has 1,458,644 examples) is treated as a stream using the first 20,000 instances, with 3 outliers removed. Similarly, from E-Power (2,075,259 examples) we use the first 100,000 instances and remove 8 outliers. Finally, from the Semi dataset (810,000 examples) we use the first 20,000 instances, retaining 19,452 after outlier removal. 
Figure~\ref{fig:imbalance_of_targets_in_datasets} illustrates target imbalance in all datasets using bin range $r=0.2$.

\paragraph{Clarification on the Integration with Python Libraries and Datasets Tested.}
In our initial submission we integrated HS and KDE with the \multiflow library. 
When we started our work, \multiflow exposed more of the internal structure of its learning algorithms, especially Hoeffding Trees, making it easier to integrate HS directly into the learning procedure. 
For example, in HAT, HS only requires the root-to-leaf prediction path, since the statistics of intermediate nodes are directly accessible regardless of additional subtrees used for drift detection.
Meanwhile, the \river library was further developed and has a more current version of iSOUP, as well as SGT which does not exist in \multiflow.  
After receiving the reviews, we integrated KDE with \river and ran additional experiments; 
however, without hyperparameter tuning due to time constraints.


Furthermore, in our initial submission we ran experiments only against three datasets: California, NY Taxi, and E-Power.  
For the camera-ready version of the paper
we decided to run experiments with Abalone too, as well as against the more challenging layers 3 and 4 of the Semi dataset.  Abalone being fairly small in size allowed us to run on time Algorithm~\ref{alg:online_ftl} on scikit-multiflow, but due to time constraints we could not do the same thing with Semi.
Experiments on layers 1 and 2 of Semi were omitted, as these cases appear trivial or easy and the KDE variants consistently outperform the baseline models.

Thus, we have experiments with \multiflow\ -testing both HS and KDE- using the tuning phases of Algorithm~\ref{alg:online_ftl} in all datasets but on Semi, while we have experiments with the \river library in all datasets -testing for potential benefit of KDE only- but without tuning this time.

Using \multiflow with Algorithm~\ref{alg:online_ftl} 
we use HT and HAT as base models. 
We equip both of these models with KDE for smoothing the imbalanced label distribution of the stream as well as HS for regularizing the predictions. These enhancements appear either together or in isolation. 
Regarding incremental KDE we use a tumbling window whose size we treat as a hyper-parameter.

\paragraph{On the Tuning Window and Hyper-Parameter Tuning.}
When running experiments with the \multiflow integration 
we use four periods for tuning the hyper-parameters of our models. 
For a dataset $S$ that we treat as a stream, we set the tuning window size to be 
$\min{\left(\left\lfloor\nicefrac{|S|}{8}\right\rfloor, 3000\right)}$.
Based on Table~\ref{tbl:datasets}, the tuning window size for each dataset tested is approximately one quarter of the ``missing'' examples from the table.  For example, in the case of Abalone, this translates to $(4,177-2,089)/4 = 522$ examples for each tuning phase.
In each tuning window 
we choose from the following hyperparameter values:
    \emph{(i)}
    for range of bin $r \in \{0, 0.1, 0.2, 0.5, 1\}$, where $0$ 
    means that we don't perform binning in our targets but every target is simply its own bin, 
    \emph{(ii)}
    $\lambda \in \{0, 0.1, 1, 10, 15, 25\}$, where $0$ is for not predicting with HS but rather with the initial unregularized model,
    \emph{(iii)}
    $h \in \{10, 50, 100\}$, 
    \emph{(iv)}
    for (tumbling) window size $\abs{W} \in \{50, 100, 200\}$, and
    \emph{(v)}
    for kernel choose between Gaussian and Epanechnikov.
When we use KDE alone we do not tune on $\lambda$, while when we use HS alone we only tune on $\lambda$. 

\paragraph{Evaluation Metrics.}
For evaluating our experiments we use:
\emph{mean absolute error (MAE)}, \emph{root mean squared error (RMSE)}, \emph{weighted root mean squared error (WRMSE)}, \emph{inverse weighted root mean squared error (INVWRMSE)}, and the \emph{$R^2$} statistic.

Let $\mathrm{bin}(y_i)$ return the bin where label $y_i$ is hashed.
For the WRMSE the penalty calculated on a set of test points \sample is 
$\mathrm{WRMSE}(f, \sample) = \sqrt{\sum_{i=1}^{\abs{S}} w(\mathrm{bin}(y_i))(f(x_i) - y_i)^2}$, where $w(\mathrm{bin}(y_i))$ is the weight of the bin where the true label $y_i$ is hashed (that corresponds to the input $x_i$) and this weight is the same for all labels that fall into the same bin.
In particular, this weight $w(\mathrm{bin}(y_i))$ corresponds to the probability of the bin where $y_i$ falls, as shown in Figure~\ref{fig:imbalance_of_targets_in_datasets} (after normalization so that we have a probability distribution); i.e., 
for some label $y$, the weight 
$w(\mathrm{bin}(y))$ is obtained from the whole distribution using bins of range $r = 0.2$, which is the range that was used to create the plots in~Figure~\ref{fig:imbalance_of_targets_in_datasets}.
The idea for WRMSE is that it assigns larger weight to instances that have more frequent labels and therefore serves as a sanity check that the learned model performs well on instances that are frequently asked.
Notice that the number of bins is in principle much smaller than the number of examples $\abs{\sample}$ and hence the weight $w(\mathrm{bin}(y_i))$ will be much larger than $1/\abs{\sample}$ for the non-empty bins that contribute in the RMSE value. For example, if we have bins with weight values $0.1$ or larger, then the $\mathrm{WRMSE}$ will be several orders of magnitude larger than simple RMSE for a datastream that has processed $n = 10,000$ examples and therefore every example in this stream contributes to RMSE a value with weight equal to $w(y_i) = 1/10,000$.
In the results that follow we will see this phenomenon. 
For the selection of the best hyper-parameters during tuning, we use the (non-weighted) RMSE.

Having said the above, in imbalanced settings it is also critical to use a metric that assigns large weights not on the high-frequency bins as WRMSE does, but rather on the low frequency bins.
Again we let $\mathrm{bin}(y_i)$ return the bin where label $y_i$ is hashed.
Hence, we define
$\mathrm{INVWRMSE}(f, \sample) = \sqrt{\sum_{i=1}^{\abs{S}} w_{\textsc{inv}}(\mathrm{bin}(y_i))(f(x_i) - y_i)^2}$, where $w_{\textsc{inv}}(\mathrm{bin}(y_i))$ is the modified weight on the bin where $y_i$ falls and has the property of being larger for bins with fewer labels (as this was obtained using bin range of size $r=0.2$).  The initial idea is to use for bin $b_j$ the weight $w'(b_j) = 1 - w(b_j)$, where $w(b_j)$ is the weight of bin $b_j$ as explained above for WRMSE.
However, we need to divide with $\sum w'(b_\ell)$ so that we can have a probability distribution with the weights. 
Therefore, for each bin $b_j$ we have
$w_{\text{inv}}(b_j) = \frac{w'(b_j)}{\sum_{\text{bins} \ b_\ell} w'(b_\ell)} = \frac{1 - w(b_j)}{\sum_{\text{bins} \ b_\ell} (1-w(b_\ell))}$, which corresponds to the weights that are used inside the $\mathrm{INVWRMSE}$ function. Finally, INVWRMSE is used once again as an additional evaluation metric; the selection of the best hyper-parameter during tuning happens by using the non-weighted RMSE.

We want to conclude our discussion on the metrics regarding a corner case that may arise.
Namely, once binning has been defined (e.g., using range $r = 0.2$), it could be the case that none of the examples from the entire distribution falls into a particular bin.
In such cases, when no examples fall into a particular bin, we observe the following.
Such empty bins will provide weight for each label $y_i$ in the bin equal to $w(\mathrm{bin}(y_i))=0$. 
This does not cause an issue in WRMSE since no instance would have true label $y_i$ that would correspond to weight $w(\mathrm{bin}(y_i)) = 0$ which is used in the formula for WRMSE.  In other words, it is never the case that such a zero weight is used in the calculation of WRMSE. 
However, for the case of INVWRMSE for such labels $y_i$ that correspond to empty bins we have $w'(\mathrm{bin}(y_i)) = 1 - w(\mathrm{bin}(y_i)) = 1 - 0 = 1$ and therefore $w_{\textsc{inv}}(\mathrm{bin}(y_i)) = \frac{w'(\mathrm{bin}(y_i))}{\sum_{bins \ b_\ell} w'(b_\ell)}$ would be positive but this value of $w'(\mathrm{bin}(y))$ would never be used immediately in some calculation, other than contributing to the total sum of $w_{\textsc{inv}}$'s in the denominator so that we can have a weight distribution where all the weights $w_{\textsc{inv}}(b_\ell)$ add up to 1 across all the bins $b_\ell$. 
%
In a sense these weights are the largest under INVWRMSE because they correspond to bins that have the most rare values $y$; i.e., those that have frequency of zero.
%
Hence, rare labels get higher importance than frequent labels, which is the point of the INVWRMSE metric.

\paragraph{Experimental Results on \multiflow.}
Table~\ref{tbl:metrics-all} presents the overall results obtained using the \multiflow library. For this reason, we report results only for HT and HAT, along with their HS, KDE, and KDE+HS extensions. 
Our results in Table~\ref{tbl:metrics-all} indicate that in all datasets except in E-Power our KDE variants (potentially with HS as well) outperform their original versions. 


\begin{table}[t]
\centering
\caption{Performance comparison across datasets on the \multiflow library using Algorithm~\ref{alg:online_ftl} for tuning. 
\textbf{\textcolor{blue}{Bold blue}} indicates best metric value per dataset, whereas plain \textbf{bold} indicates the second best. For comparison purposes on plain HT and HAT we bypass the tuning portions of the stream.}
\label{tbl:metrics-all}
\resizebox{\textwidth}{!}{
\begin{tabular}{llrrrrrrrr}
\toprule
%
 \multicolumn{2}{c}{} & \multicolumn{8}{c}{\textbf{\scriptsize{Model (potentially including HS, KDE, or both)}}}
\\
\cmidrule{3-10}
 & & & & &  &  &  & \textbf{\scriptsize{HT}} & \textbf{\scriptsize{HAT}} 
 \\
 & & & & 
 \textbf{\scriptsize{HT}} & \textbf{\scriptsize{HAT}} & \textbf{\scriptsize{HT}} & \textbf{\scriptsize{HAT}} & \textbf{\scriptsize{+HS}} & \textbf{\scriptsize{+HS}} \\
\textbf{Dataset} & \textbf{Metric} 
& \rotatebox{0}{\textbf{\scriptsize{HT}}}
& \rotatebox{0}{\textbf{\scriptsize{HAT}}}
& \rotatebox{0}{\textbf{\scriptsize{+HS}}}
& \rotatebox{0}{\textbf{\scriptsize{+HS}}}
& \rotatebox{0}{\textbf{\scriptsize{+KDE}}}
& \rotatebox{0}{\textbf{\scriptsize{+KDE}}}
& \rotatebox{0}{\textbf{\scriptsize{+KDE}}}
& \rotatebox{0}{\textbf{\scriptsize{+KDE}}} \\
\midrule

\multirow{5}{*}{Abalone}
& MAE  
& 1.979 & 1.979 & 1.979 & 1.979 
& \textbf{1.643} & 1.648 
& 1.644 
& \textbf{\textcolor{blue}{1.560}} \\

& RMSE 
& 2.634 & 2.634 & 2.634 & 2.634 
& 2.350 & 2.350 
& \textbf{2.347} 
& \textbf{\textcolor{blue}{2.233}} \\

& WRMSE 
& \textbf{7.037} & \textbf{7.037} & \textbf{\textcolor{blue}{6.916}}  & \textbf{\textcolor{blue}{6.916}}
& 7.255 & 7.508 
& 7.210
& 7.471  \\

& INVWRMSE
& \textbf{4.736} & \textbf{4.736} & \textbf{\textcolor{blue}{4.733}}
& \textbf{\textcolor{blue}{4.733}} & 4.832 & 4.939
& 4.824 & 4.930
 \\

& $R^2$ 
& -0.015 & -0.015 & -0.015 & -0.015 
& 0.192 & 0.192 
& \textbf{0.194} 
& \textbf{\textcolor{blue}{0.271}} \\

\midrule

\multirow{5}{*}{California}
& MAE  
& 0.668 & 0.692 & 0.659 & 0.700 
& \textbf{0.568} & 0.625 
& \textbf{\textcolor{blue}{0.564}} 
& 0.628 \\

& RMSE 
& 0.904 & 0.931 & 0.899 & 0.940 
& \textbf{0.808} & 0.914 
& \textbf{\textcolor{blue}{0.802}} 
& 0.915 \\

& WRMSE 
& 20.282 & 21.793 & 19.755 & 21.966
& \textbf{18.851} & 22.148
& \textbf{\textcolor{blue}{18.557}} 
& 22.227  \\

& INVWRMSE
& 18.286 &
18.789 &
18.202 &
18.962 &
\textbf{16.309} &
18.401 &
\textbf{\textcolor{blue}{16.191}} &
18.413 \\

& $R^2$ 
& 0.380 & 0.342 & 0.387 & 0.330 
& \textbf{0.505} & 0.367 
& \textbf{\textcolor{blue}{0.512}} 
& 0.366 \\

\midrule

\multirow{5}{*}{NY Taxi}
& MAE  
& 0.466 & 0.466 & 0.466 & 0.466 
& \textbf{\textcolor{blue}{0.368}} 
& \textbf{\textcolor{blue}{0.368}} 
& \textbf{0.369} 
& \textbf{0.369} \\

& RMSE 
& 0.612 & 0.612 & 0.612 & 0.612 
& \textbf{\textcolor{blue}{0.487}} 
& \textbf{\textcolor{blue}{0.487}} 
& \textbf{0.487} 
& \textbf{0.487} \\

& WRMSE 
& 13.177 & 13.177 & 13.147 & 13.147
& \textbf{11.932} & \textbf{11.932} 
& \textbf{\textcolor{blue}{11.786}} 
& \textbf{\textcolor{blue}{11.786}}  \\

& INVWRMSE
& 12.454 & 12.454 & 12.454
& 12.454 & \textbf{\textcolor{blue}{9.841}} & \textbf{\textcolor{blue}{9.841}}
& \textbf{9.855} & \textbf{9.855}  \\

& $R^2$ 
& 0.333 & 0.333 & 0.333 & 0.333 
& \textbf{\textcolor{blue}{0.577}} 
& \textbf{\textcolor{blue}{0.577}} 
& \textbf{0.577} 
& \textbf{0.577} \\

\midrule

\multirow{5}{*}{E-Power}
& MAE  
& \textbf{\textcolor{blue}{0.080}} 
& 0.407 
& \textbf{0.098} 
& 0.419 
& 0.114 
& 0.229 
& 0.114 
& 0.230 \\

& RMSE 
& \textbf{\textcolor{blue}{0.203}} 
& 0.660 
& \textbf{0.219} 
& 0.666 
& 0.272 
& 0.479 
& 0.272 
& 0.479 \\

& WRMSE 
& \textbf{\textcolor{blue}{9.460}} & 41.994 & \textbf{12.371} & 43.223 
& 15.299 & 29.430 
& 15.310
& 29.479 \\

& INVWRMSE
& \textbf{\textcolor{blue}{8.765}} & 28.201 & \textbf{9.399}
& 28.441 & 11.679 & 20.481
& 11.678 & 20.484 \\

& $R^2$ 
& \textbf{\textcolor{blue}{0.977}} 
& 0.756 
& \textbf{0.973} 
& 0.752 
& 0.959 
& 0.872 
& 0.959 
& 0.872 \\

\bottomrule
\end{tabular}
}
\end{table}

\begin{figure}[!ht]
\begin{subfigure}[b]{0.5\columnwidth}
\centering
\includegraphics[width=\columnwidth]{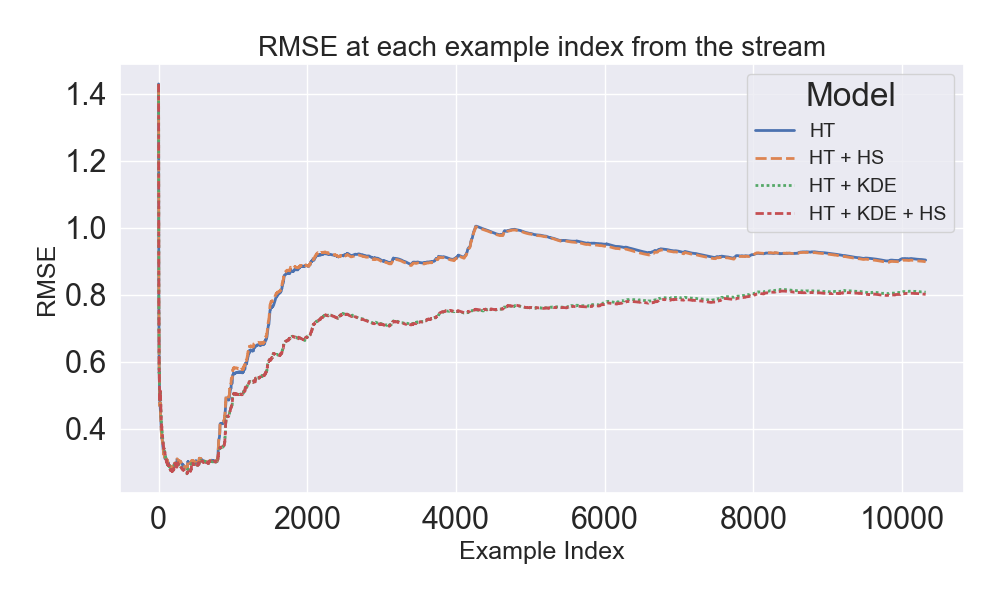}
\caption{Cumulative RMSE.}
\label{fig:california_hat_rmse}
\end{subfigure}
\hfill
\begin{subfigure}[b]{0.5\columnwidth}
\centering
\includegraphics[width=\columnwidth]{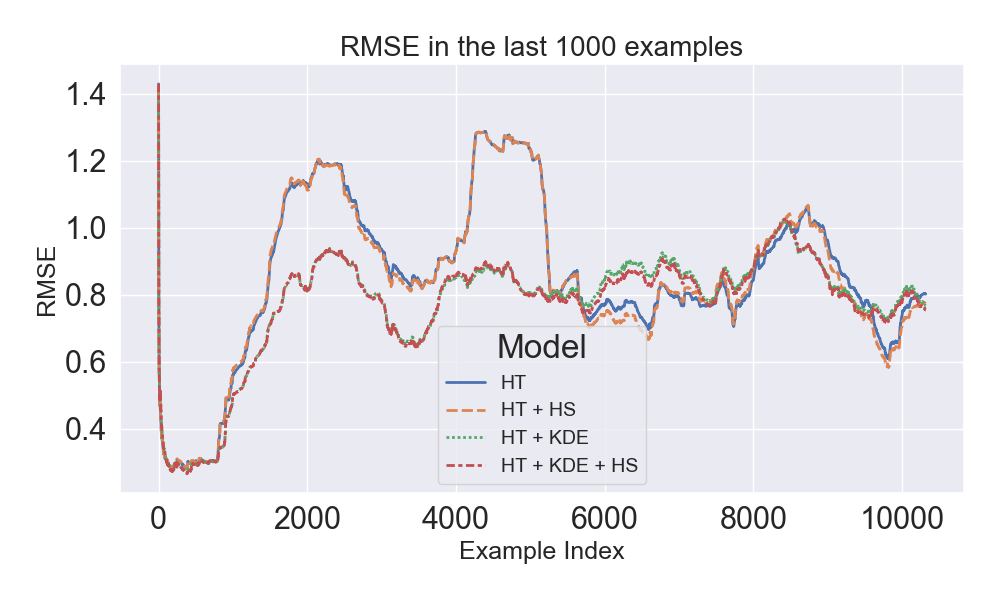}
\caption{RMSE in the last 1K examples.}
\label{fig:california_hat_rmse_1K}
\end{subfigure}
\caption{RMSE plots of HT-based models for the \emph{California} dataset. The KDE and KDE+HS variants perform overall better than the original HT model. Experiments performed on \multiflow library.}
\label{fig:california_hat_rmse_total}
\end{figure}

\begin{figure}[!ht]
\begin{subfigure}[b]{0.5\columnwidth}
\centering
\includegraphics[width=\columnwidth]{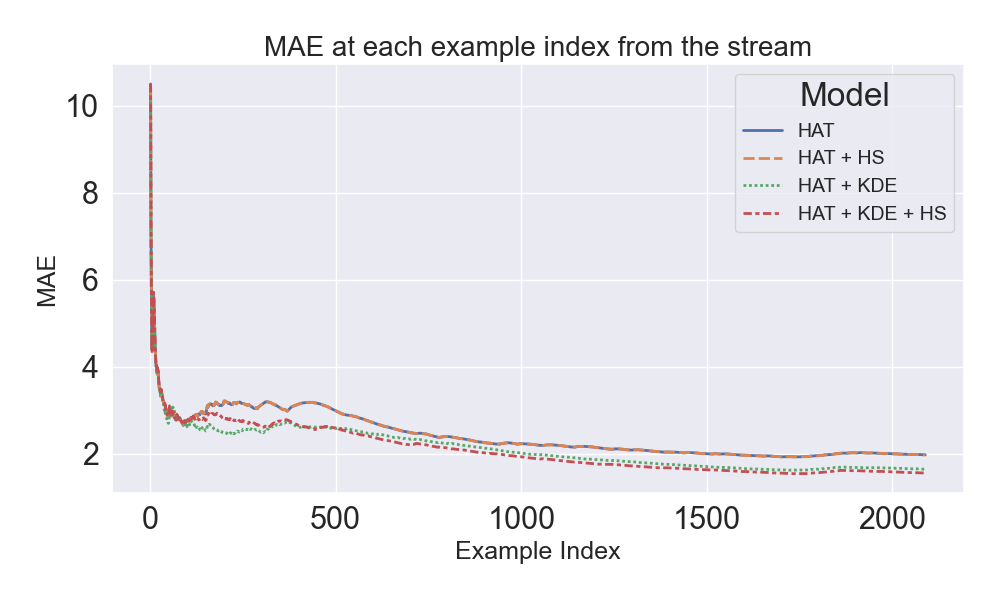}
\caption{Cumulative MAE.}
\label{fig:abalone_hat_mae}
\end{subfigure}
\hfill
\begin{subfigure}[b]{0.5\columnwidth}
\centering
\includegraphics[width=\columnwidth]{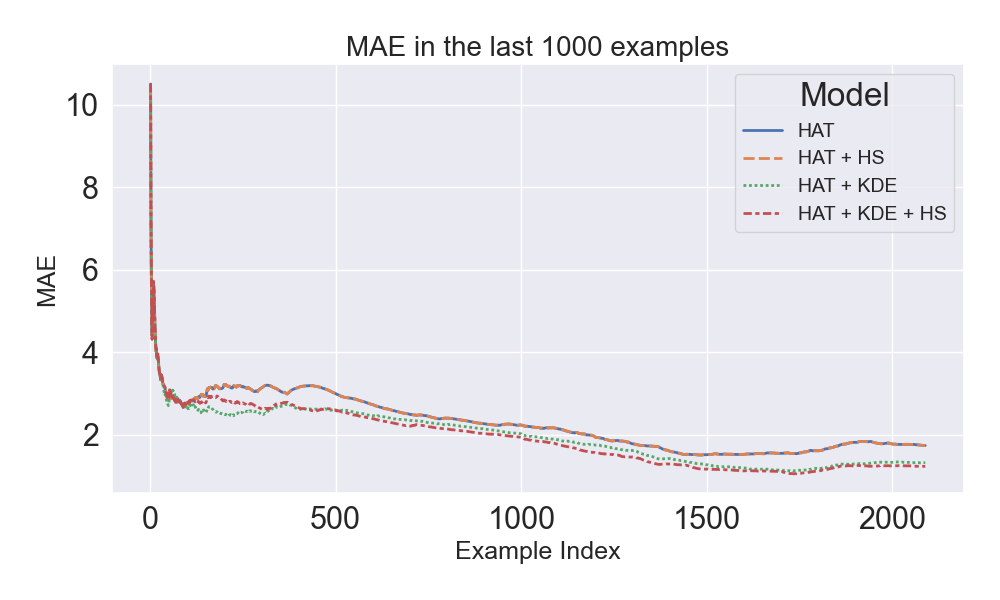}
\caption{MAE in the last 1K examples.}
\label{fig:abalone_hat_mae_1K}
\end{subfigure}
\caption{MAE plots of HAT-based models for the \emph{Abalone} dataset. Similar to Figure~\ref{fig:california_hat_rmse_total} the KDE-enhanced model performs better than the original model. However, here HS helps even further.
Experiments performed on \multiflow library.}
\label{fig:abalone_hat_mae_total}
\end{figure}

\begin{figure}[!ht]
\begin{subfigure}[b]{0.5\columnwidth}
\centering
\includegraphics[width=\columnwidth]{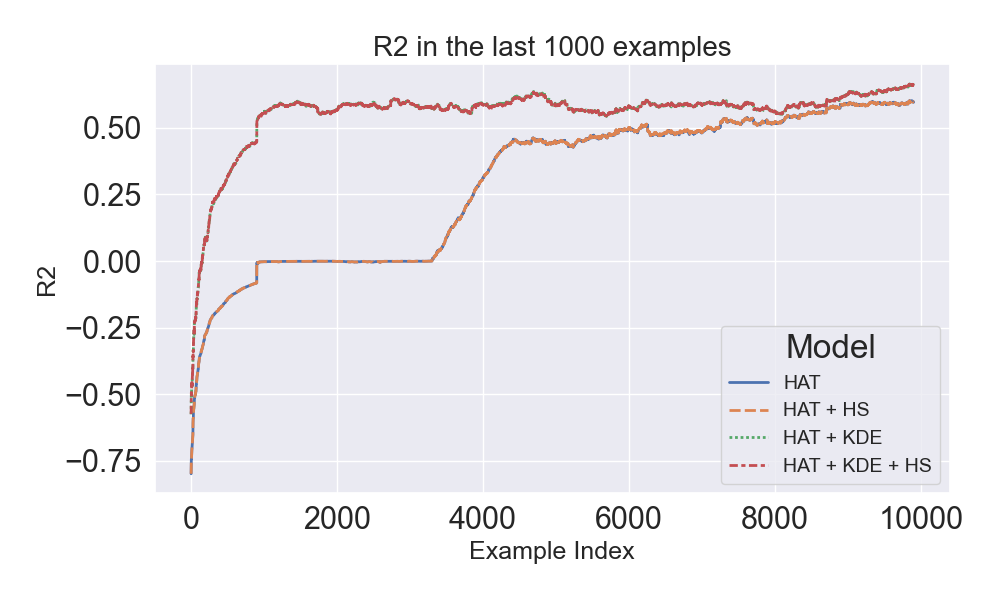}
\caption{$R^2$ in the last 1K examples (NY Taxi).}
\label{fig:nytaxi_hat_r2_1K}
\end{subfigure}
\hfill
\begin{subfigure}[b]{0.5\columnwidth}
\centering
\includegraphics[width=\columnwidth]{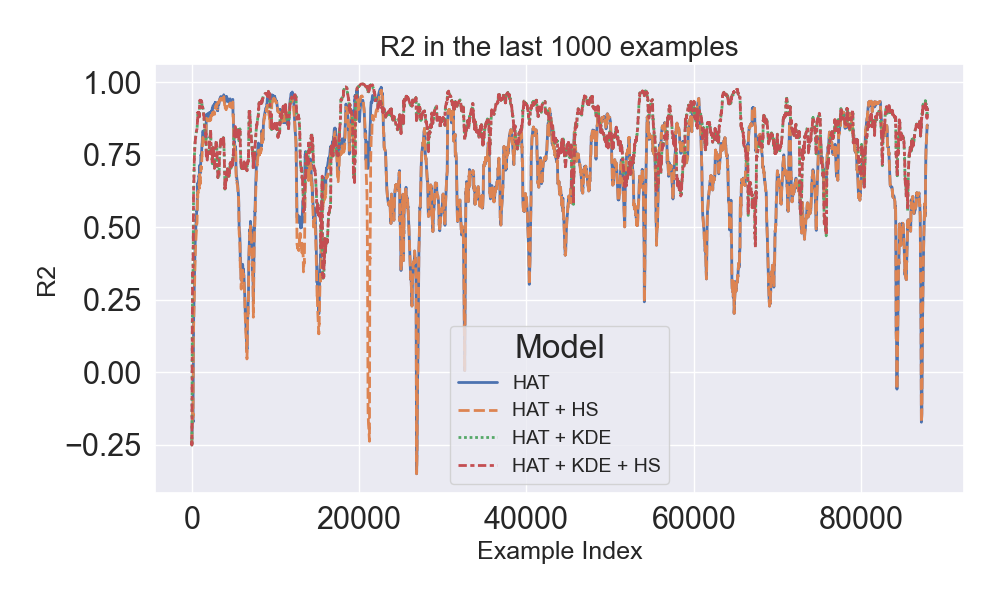}
\caption{$R^2$ in the last 1K examples (E-Power).}
\label{fig:epower_hat_r2_1K}
\end{subfigure}
\caption{$R^2$ score during the last 1K examples for HAT-based models on the NY Taxi (Fig.~\ref{fig:nytaxi_hat_r2_1K})  and E-Power (Fig.~\ref{fig:epower_hat_r2_1K}) datasets. In both cases, the KDE variants achieve the best performance throughout the streaming process. 
Furthermore, in 
both cases 
plotting starts after 100 examples have first been processed since the early $R^2$ values are very small and dominate the y-axis in the graph.
Experiments performed on \multiflow library.}
\label{fig:nytaxi_epower_r2_1K}
\end{figure}

We now provide some additional information complementing the results of Table~\ref{tbl:metrics-all}.
In 
Figure~\ref{fig:california_hat_rmse_total} we see that for the California dataset, the performance of the KDE and the KDE+HS variants of HT outperform the original model, with the largest performance gap appearing in the middle of the stream. 
In Figure~\ref{fig:abalone_hat_mae_total}
we see the MAE performance of HAT models on the Abalone dataset.
This time the KDE+HS enhanced version of HAT behaves the best and is discernibly better compared to the simple KDE extension of the HAT model.
In Figure~\ref{fig:nytaxi_epower_r2_1K} we see two graphs of $R^2$ for the NY Taxi and the E-Power datasets using again HAT-based models for comparisons. In both cases we see that the KDE-enhanced and KDE+HS-enhanced variants perform better than the base model. Having said that, for the case of E-Power, the simple HT performs even better (see Table~\ref{tbl:metrics-all}).
%


\paragraph{Experimental Results on \river.}
Following the reviewers’ feedback, we executed additional experiments using the \river Python library as well. 
This allowed us to evaluate our KDE framework with the iSOUP and SGT models beyond HT and HAT that we evaluated using \multiflow. 
Due to time constraints, we did not perform online tuning using Algorithm~\ref{alg:online_ftl}. 
Instead, we only performed incremental KDE as described in Algorithm~\ref{alg:incr_kde}, using the same fixed parameters ($h=10$, $r=0.2$ and window size $\abs{W}=50$) for all the experiments.

\begin{figure}[!t]
\begin{subfigure}[b]{0.5\columnwidth}
\centering
\includegraphics[width=\columnwidth]{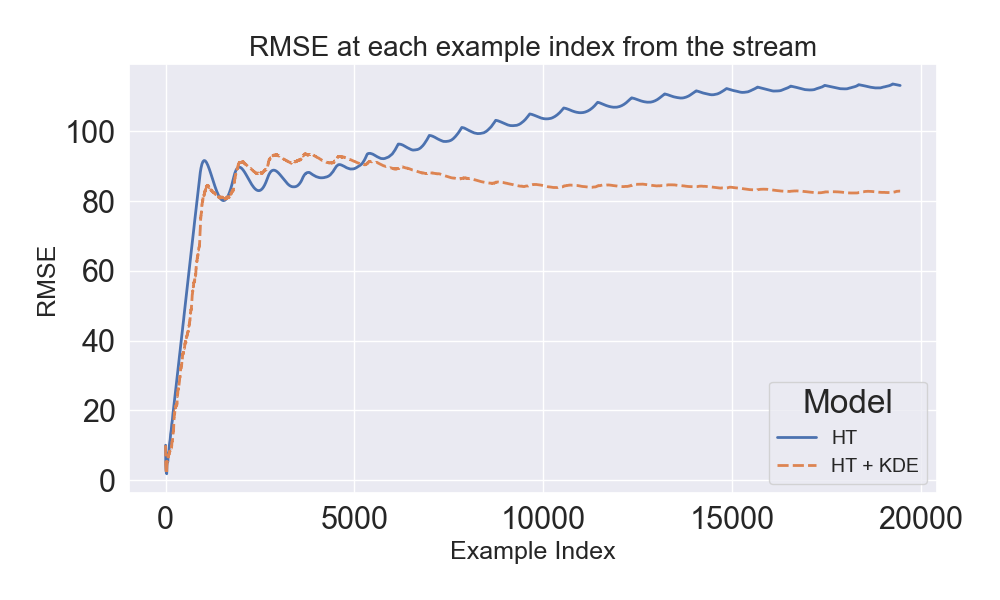}
\caption{Cumulative RMSE in Semi-L3.}
\label{fig:l3_cumulative_rmse}
\end{subfigure}
\hfill
\begin{subfigure}[b]{0.5\columnwidth}
\centering
\includegraphics[width=\columnwidth]{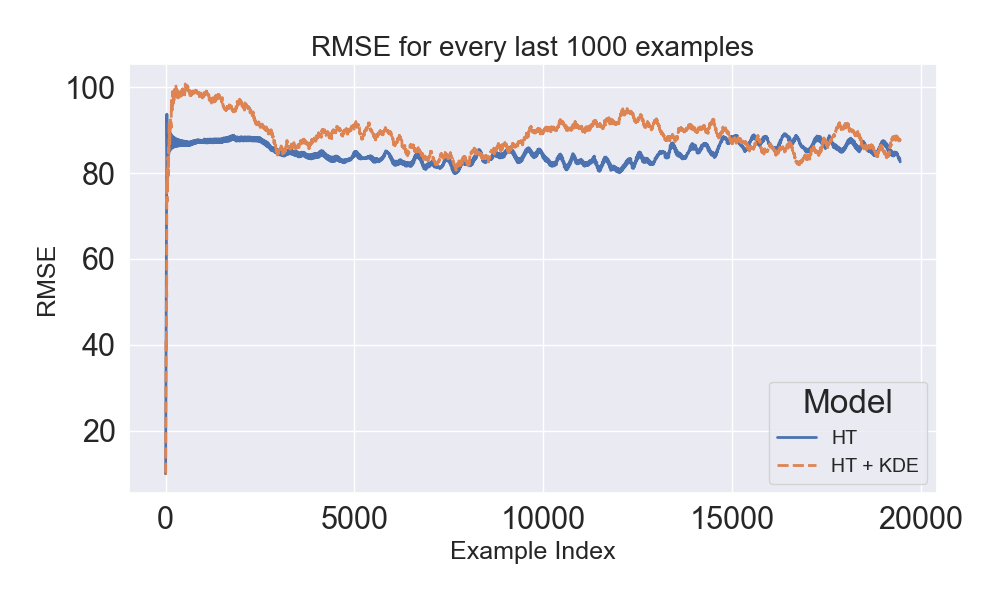}
\caption{RMSE for 1K last examples in Semi-L4.}
\label{fig:l4_rmse_1K}
\end{subfigure}
\caption{RMSE, cumulative and during the last 1K examples, for HT-based models on the Semi-L3 (Fig.~\ref{fig:l3_cumulative_rmse}) and Semi-L4 (Fig.~\ref{fig:l4_rmse_1K}) datasets. 
Note that these results were obtained using the \river library.}
\label{fig:l3_cumulative_rmse_l4_1K_rmse}
\end{figure}

Table~\ref{tbl:river_all_datasets} presents results with the \river library, 
using all five metrics that we used for the \multiflow library too.
We denote layers 3 and 4 of the Semi dataset as Semi-L3 and Semi-L4 respectively. 
As we had no plots regarding the Semi dataset generated using models from \multiflow, in Figure~\ref{fig:l3_cumulative_rmse_l4_1K_rmse} we now plot two graphs of the Semi dataset focusing on layers 3 and 4. Regarding performance in layer 3 in Figure~\ref{fig:l3_cumulative_rmse} HT performs better using KDE, but in Figure~\ref{fig:l4_rmse_1K} the KDE variant is underperforming during most of the streaming process and becomes equivalent to plain HT near the end.

Coming back to Table~\ref{tbl:river_all_datasets}, 
a first remark is that KDE is in general helpful for the base models.  
In the case of MAE, the KDE-enhanced models have better performance in 17 out of 24 cases.
In the case of RMSE and  INVWRMSE, the KDE-enhanced models have better performance in 18 out of 24 cases, while in the case of WRMSE the KDE-enhanced models have better performance in 15 out of 24 cases.
Regarding $R^2$, the KDE-enhanced models have better performance in 16 out of 24 cases.
Overall, 
depending on the performance metric, MAE, RMSE, WRMSE, INVWRMSE, and $R^2$,  
the KDE-enhanced models provide the best performing models in six out of six datasets, five out of six datasets, four out of six datasets, five out of six datasets, and five out of six datasets respectively.

\begin{table}[!t]
\centering
\caption{Performance comparison 
across datasets using the \river Python library. \textbf{Bold} values indicate the better performance between each base model and its KDE-enhanced counterpart. \textcolor{blue}{\textbf{Bold blue}} indicates the best performing model for a given dataset and metric. 
As mentioned in the text, due to time constraints we did not perform tuning to identify the best hyper-parameters in these experiments involving the \river library.
For the results shown for Semi-L3 and Semi-L4 we are using only one decimal digit for better clarity. When necessary we may take into account additional decimal digits to break seeming ties regarding the best performing model (e.g., in the case of Semi-L4, plain $R^2$ for HT and SGT+KDE appear to be both $0.0$, but HT is actually higher).
}
\label{tbl:river_all_datasets}
\resizebox{\textwidth}{!}{
\begin{tabular}{llrrrrrrrr}
\toprule
\multicolumn{2}{c}{} & \multicolumn{8}{c}{\textbf{Model (potentially including KDE)}}
\\
\cmidrule{3-10}
 & & & & & & \textbf{HT}
& \textbf{HAT}
& \textbf{iSOUP}
& \textbf{SGT} \\
\textbf{Dataset} & \textbf{Metric}
& \textbf{HT}
& \textbf{HAT}
& \textbf{iSOUP}
& \textbf{SGT}
& \textbf{+KDE}
& \textbf{+KDE}
& \textbf{+KDE}
& \textbf{+KDE} \\
\midrule

\multirow{5}{*}{Abalone}
& MAE
& 2.351 
& 2.357 
& \textbf{10.802} 
& 2.622
& \textbf{2.052}
& \textbf{\textcolor{blue}{1.863}} 
& 10.864
& \textbf{2.166}
\\
& RMSE
& 3.175 & 3.186 & \textbf{11.249} & 4.051
& \textbf{2.776} & \textcolor{blue}{\textbf{2.633}} & 11.318 & \textbf{3.138} \\
& WRMSE
& \textbf{43.815} & \textcolor{blue}{\textbf{42.957}} & \textbf{215.832} & 64.202
& 50.971 & 46.795 & 218.941 & \textbf{57.864} 
\\
& INVWRMSE
& 16.944
& 17.018
& \textbf{58.676}
& 21.451
& \textbf{14.539}
& \textbf{\textcolor{blue}{13.826}}
& 58.988
& \textbf{16.430}
\\
& $R^2$
& 0.030
& 0.023
& \textbf{-11.176}
& -0.579
& \textbf{0.258}
& \textbf{\textcolor{blue}{0.333}}
& -11.326
& \textbf{0.052} \\
\midrule

\multirow{5}{*}{California}
& MAE
& 0.634
& 0.608
& \textbf{2.067}
& 0.786
& \textbf{0.552} 
& \textbf{\textcolor{blue}{0.536 }}
& \textbf{2.067}
& \textbf{0.702} \\

& RMSE
& 0.837 & 0.827 & 2.367 & 1.033
& \textcolor{blue}{\textbf{0.768}} & \textbf{0.770} & \textbf{2.366} & \textbf{0.940} \\
& WRMSE
& 26.069 & 26.725 & 70.863 & 32.971
& \textcolor{blue}{\textbf{24.118}} & \textbf{24.753} & \textbf{70.852} & \textbf{29.310} \\
& INVWRMSE
& 23.949
& 23.614
& 67.876
& 29.540
& \textbf{\textcolor{blue}{21.966}}
& \textbf{22.011}
& \textbf{67.865}
& \textbf{26.912} 
\\
& $R^2$
& 0.474 
& 0.487 
& \textbf{-3.207}
& 0.198 
& \textbf{\textcolor{blue}{0.557}} 
& \textbf{0.555}

& \textbf{-3.207}

& \textbf{0.336} 
\\

\midrule

\multirow{5}{*}{NY Taxi}
& MAE
& 0.419 
& 0.394 
& 6.450 
& 0.447 
& \textbf{0.363}
& \textbf{0.369} 
& \textbf{6.444} 
& \textbf{\textcolor{blue}{0.360}} 
\\
& RMSE
& 0.553 & 0.525 & 6.494 & 0.829
& \textcolor{blue}{\textbf{0.485}} & \textbf{0.491} & \textbf{6.489} & \textbf{0.511} \\
& WRMSE
& 17.494 & \textbf{16.797} & 254.611 & 30.739
& \textcolor{blue}{\textbf{16.664}} & 16.981 & \textbf{254.379} & \textbf{17.625} \\
& INVWRMSE
& 15.901
& 15.080
& 183.987 
& 23.600 
& \textbf{\textcolor{blue}{13.870}}
& \textbf{14.052}
& \textbf{183.831} 
& \textbf{14.621}
\\
& $R^2$
& 0.457 
& 0.511 
& -73.812 
& -0.220 
& \textbf{\textcolor{blue}{0.583}}
& \textbf{0.572}
& \textbf{-73.684} 
& \textbf{0.536} 
\\
\midrule

\multirow{5}{*}{E-Power}
& MAE
& \textbf{0.077}
& 0.442 
& 1.646
&\textbf{0.296}
& \textbf{\textcolor{blue}{0.037}}
& \textbf{0.189}
& \textbf{1.645} 
& 0.423 
\\
& RMSE
& 0.160 & 0.672 & 2.124 & \textbf{0.477}
& \textcolor{blue}{\textbf{0.082}} & \textbf{0.403} & \textbf{2.124} & 0.574 \\
& WRMSE
& 11.136 & 50.203 & 112.694 & \textbf{38.406}
& \textcolor{blue}{\textbf{4.074}} & \textbf{25.534} & \textbf{112.657} & 52.591 
\\
& INVWRMSE
& 7.268 
& 30.452
& 97.606
& \textbf{21.493} 
& \textbf{\textcolor{blue}{3.767}}
& \textbf{18.415} 
& \textbf{97.602}
& 25.594
\\
& $R^2$
& 0.986
& 0.750
& \textbf{-1.497} 
& \textbf{0.874} 
& \textbf{\textcolor{blue}{0.996}} 
& \textbf{0.910}
& \textbf{-1.497} 
& 0.818 \\
\midrule

\multirow{5}{*}{Semi-L3}
& MAE
& 97.3 
& 66.9 
& \textbf{151.6}
& \textbf{70.8} 
& \textbf{62.6}
& \textbf{\textcolor{blue}{27.5}} 
& 152.7 
& 73.4 
\\
& RMSE
& 113.1 
& 80.3 
& \textbf{174.8} 
& \textbf{85.2} 
& \textbf{82.8} 
& \textcolor{blue}{\textbf{55.4}} 
& 175.9
& 91.3 
\\
& WRMSE
& 2866.6 
& 2058.1 
& \textbf{4421.1} 
& \textbf{2174.6} 
& \textbf{2114.7} 
& \textcolor{blue}{\textbf{1432.0}} 
& 4449.2
& 2339.2
\\
& INVWRMSE
& 407.5
& 289.0
& \textbf{629.6} 
& \textbf{306.8} 
& \textbf{298.3}
& \textbf{\textcolor{blue}{199.3}}
& 633.6
& 328.8
\\
& $R^2$
& -0.7 
& 0.1
& \textbf{-3.0}
& \textbf{0.0}
& \textbf{0.0}
& \textbf{\textcolor{blue}{0.6}}
& -3.1
& -0.1
\\
\midrule

\multirow{5}{*}{Semi-L4}
& MAE
& 70.3
& 72.9
& 153.4
& \textbf{68.9} 
& \textbf{\textcolor{blue}{67.1}}
& \textbf{74.1} 
& \textbf{153.3} 
& 74.4
\\
& RMSE
& \textcolor{blue}{\textbf{84.7}} 
& \textbf{85.4}
& 176.2
& 88.7
& 88.9
& 95.4
& \textbf{176.0} 
& \textbf{88.2}
\\
& WRMSE
& \textcolor{blue}{\textbf{2158.4}} 
& \textbf{2177.3} 
& 4464.8
& 2261.4
& 2265.6
& 2430.4
& \textbf{4461.0}
& \textbf{2250.9}
\\
& INVWRMSE
& \textbf{\textcolor{blue}{305.0}}
& \textbf{307.5}
& 634.5
& 319.5
& 320.1
& 343.4
& \textbf{633.9}
& \textbf{317.7}
\\
& $R^2$
& \textbf{\textcolor{blue}{0.0}} 
& \textbf{0.0} 
& -3.2 
& -0.1 
& -0.1 
& -0.2 
& \textbf{-3.1} 
& \textbf{0.0} 
\\
\bottomrule
\end{tabular}
}
\end{table}

\paragraph{A Clarification on the Experimental Results.}
The reader may have noticed that the plain HT and HAT learned models attain slightly different performance in Tables~\ref{tbl:metrics-all} and~\ref{tbl:river_all_datasets} between the experiments that we have conducted with the two libraries whenever the datasets overlap. 
There can be multiple explanations for this.
First, is that while the predict-then-train method is being used to calculate the metric values in both cases, nevertheless, in the experiments with the \river library we use the entire sequence of the dataset, while in the experiments with the \multiflow library we bypass parts of the stream in our calculations as we are using these parts to tune our models.
Second, is that the weights $w(y)$ and $w_{\textsc{inv}}(y)$ that we use for the weighted metrics are different between the two experiments for the same reason.  While in both cases we see the whole train-then-predict sequence ahead of time in order to determine the weights that contribute to the appropriate metric, again, these sequences are different since there are portions of the original dataset that are used only for tuning in the case of the \multiflow library.\footnote{As a minor remark, notice that the knowledge of the whole distribution ahead of time only affects the penalty that we calculate for each learned model.  However, this information of the true distribution is never leaked into the learning process that could potentially invalidate our experimental results. As a related comment, in the KDE-enhanced versions of the learned models, we are using a tumbling window to approximate the label distribution and this is where the KDE approach is being applied to; i.e., only on examples that we have seen and in fact only on examples that fit into the tumbling window.}  
In other words, not the same distribution of frequencies is used to determine the weights that we integrate in the metrics.
Third, there can also be slight implementation differences between the two models that could result in some discrepancies despite the fact that we have otherwise tried to align the behavior of the learned models between the two libraries; e.g., we are using \texttt{mean} as the appropriate method for leaf prediction.
Finally, differences in KDE-enhanced models are also expected since in the case of \multiflow library we also perform tuning, while in the case of \river there is no tuning phase.

\paragraph{Resources Used for Experiments.}
All experiments were executed on a macOS laptop with an Apple Silicon M2 processor and 16\,GB of RAM. We implemented the models in Python using the \multiflow and \river libraries.

\section{Conclusions and Future Work}
We recast recent advances in \emph{batch learning} regarding imbalanced data and regularization of decision trees, in \emph{streaming scenarios}. 
We propose the inclusion of KDE and potentially HS in incremental decision tree learning algorithms. 
We have seen that, in general, KDE improves 
the performance of the base models. 
However, HS appears to be providing minimal gains, if at all. 
As our results improve the performance of base incremental decision trees used for regression, 
we anticipate that the observed benefits will translate to 
similar benefits 
in random forests, or other ensembles of such tree-based models.
We note that KDE does not naturally extend to pure classification problems and it is a question if one can find a new method that can further improve the performance of such incremental models with, or without, the inclusion of HS.  Furthermore, in this work we have not focused on drift, which is an exciting direction, especially when one would like to couple drift together with imbalanced data and potentially KDE.


\paragraph{Acknowledgements.}
This material is based upon work supported by the U.S.~National Science Foundation under Grant No.~RISE-2019758.  This work is part of the NSF AI Institute for Research on Trustworthy AI in Weather, Climate, and Coastal Oceanography (NSF AI2ES).

\paragraph{Disclosure of Interests.}
The authors have no competing interests to declare that are relevant to the content of this article.

\bibliographystyle{plainnat}
\bibliography{references-short}

\end{document}